\documentclass[10pt,twocolumn,letterpaper]{article}

\usepackage{cvpr}              

\usepackage[dvipsnames]{xcolor}

\definecolor{cvprblue}{rgb}{0.21,0.49,0.74}
\usepackage[pagebackref,breaklinks,colorlinks,citecolor=cvprblue]{hyperref}
\usepackage{multirow}
\usepackage{soul}
\usepackage{url}
\usepackage{color}
\usepackage{balance} 
\usepackage{xspace}
\usepackage{tabularx}
\usepackage{colortbl}
\usepackage{enumitem}
\usepackage{wasysym}
\usepackage{pifont}

\definecolor{bblue}{rgb}{0,150,230}
\definecolor{mygray}{gray}{.9}
\definecolor{lightgray}{gray}{.96}
\definecolor{myy}{RGB}{126,95,0}
\definecolor{ggray}{RGB}{127,127,127}
\definecolor{mygreen}{RGB}{93,173,85}
\definecolor{myred}{RGB}{240,16,89}
\definecolor{myblue}{RGB}{0,114,188}
\definecolor{darkgreen}{rgb}{0.0, 0.5, 0.0}
\definecolor{demphcolor}{RGB}{100,100,100}
\definecolor{sh_blue}{rgb}{0,0.60,0.93}

\def\paperID{11148} 
\def\confName{CVPR}
\def\confYear{2024}

\title{Unsupervised Low-light Image Enhancement with Lookup Tables \\and Diffusion Priors}

\author{
Yunlong Lin$^{1\scalebox{0.75}{*}}$ \quad Zhenqi Fu$^{2}$\thanks{Yunlong Lin and Zhenqi Fu contribute equally to this work.} \quad Kairun Wen$^{1}$ \quad Tian Ye$^{3}$ \quad Sixiang Chen$^{3}$\\ \quad Ge Meng$^{1}$\quad Yingying Wang$^{1}$ \quad Yue Huang$^{1}$ \quad Xiaotong Tu$^{1}$ \quad Xinghao Ding$^{1}$\thanks{Xinghao Ding (dxh@xmu.edu.cn) is the corresponding author.}\\ \vspace{-0.5mm}
  $^{1}$Xiamen University\quad
  $^{2}$Tsinghua University\quad \\
  $^{3}$The Hong Kong University of Science and Technology (Guangzhou)\quad \\ \vspace{-0.5mm}
{\tt\small Project page: \url{https://dplut.github.io/}}
}

\begin{document}
 \maketitle


 \begin{abstract}
Low-light image enhancement (LIE) aims at precisely and efficiently recovering an image degraded in poor illumination environments. Recent advanced LIE techniques are using deep neural networks, which require lots of low-normal light image pairs, network parameters, and computational resources. As a result, their practicality is limited. In this work, we devise a novel unsupervised LIE framework based on \textbf{\underline{d}}iffusion \textbf{\underline{p}}riors and lookup tables (DPLUT) to achieve efficient low-light image recovery. The proposed approach comprises two critical components: a \textbf{\underline{l}}ight adjustment lookup table (LLUT) and a \textbf{\underline{n}}oise suppression lookup table (NLUT). LLUT is optimized with a set of unsupervised losses. It aims at predicting pixel-wise curve parameters for the dynamic range adjustment of a specific image. NLUT is designed to remove the amplified noise after the light brightens. As diffusion models are sensitive to noise, diffusion priors are introduced to achieve high-performance noise suppression. Extensive experiments demonstrate that our approach outperforms state-of-the-art methods in terms of visual quality and efficiency.
\end{abstract}    
 \section{Introduction}

The goal of low-light image enhancement (LIE) is to improve the visual quality of images captured in low-light conditions. As a fundamental preprocessing task, LIE algorithms are expected to be effective and efficient, especially on resource-constrained devices and embedded platforms. Over the past few years, prolific algorithms have been proposed, which can be roughly classified into efficiency- and quality-oriented methods.

\begin{figure}[t]
	\centering
 \setlength{\abovecaptionskip}{0.1cm} 
    \setlength{\belowcaptionskip}{-0.1cm}
	\includegraphics[width=1\linewidth]{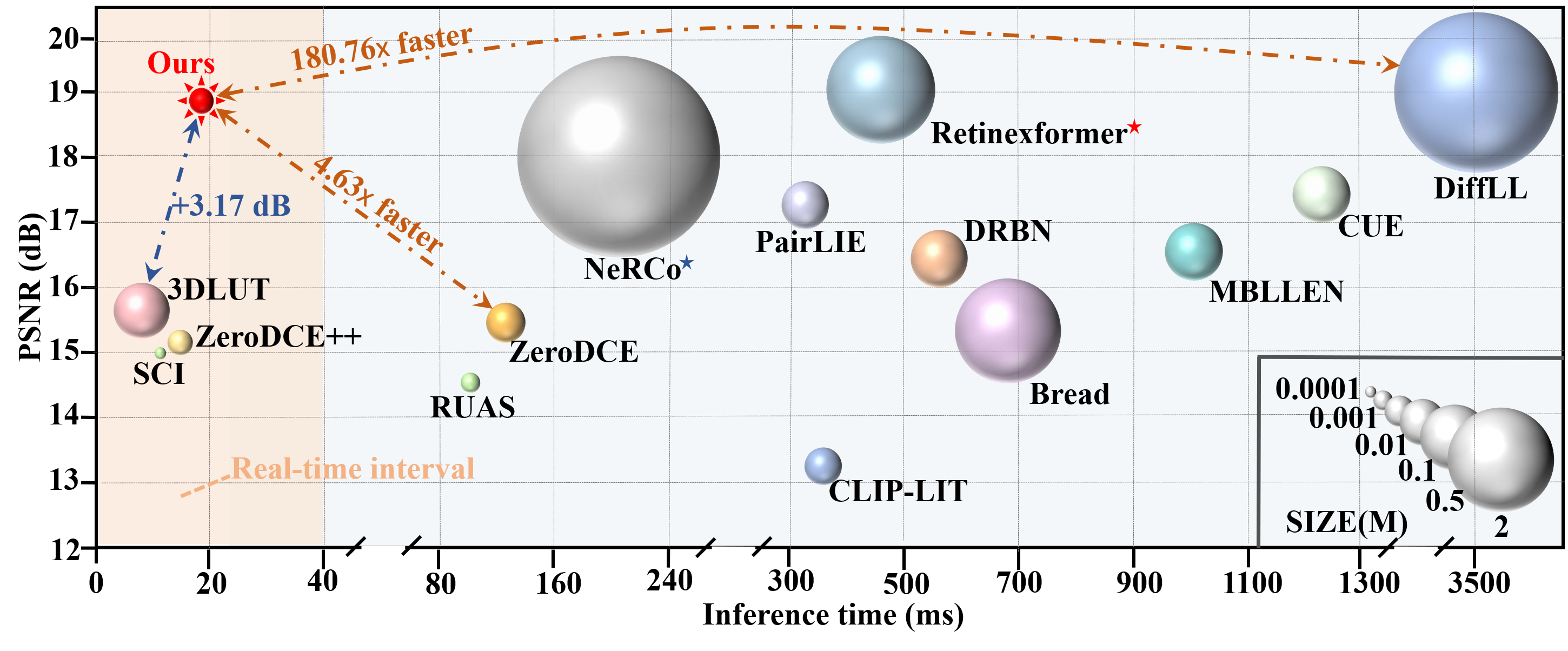}
\caption{\textbf{Comparisons of performance and efficiency. }The average PSNR is evaluated on LSRW~\cite{LSRW}, and inference time is evaluated on 4K ($3840\times 2160$) resolution with a single Titan RTX GPU. Our approach obtains the highest PSNR and can process 4K low-light images in real-time. Note that \textcolor{blue}{$^{\star}$} and \textcolor{red}{$^{\star}$} indicate the maximum size that the models can handle is 480P ($640\times 480$) and 1080P ($1920\times 1080$), respectively.}\label{intro}
\end{figure}

Efficiency-oriented methods, such as SCI~\cite{SCI} and ZeroDCE++~\cite{Zerodcepp}, can generate normal-light images in real-time. However, these methods have limited representative ability in diverse low-light degradation, resulting in suboptimal performance. As shown in Fig.~\ref{intro}, efficiency-oriented methods can meet the real-time requirements of practical applications, but the performance is far from satisfactory. In contrast, quality-oriented approaches, such as Retinexformer~\cite{Retinexformer} and CUE~\cite{CUE} achieve relatively higher performance than efficiency-oriented methods. Their success was largely due to the deep and complex network architectures, massive paired training data, and high computational and memory costs. As presented in Fig.~\ref{intro}, existing quality-oriented approaches cannot process 4K resolution images in real-time.
\begin{figure*}[t]
	\centering
 \setlength{\abovecaptionskip}{0.1cm} 
    \setlength{\belowcaptionskip}{-0.1cm}
	\includegraphics[width=1\linewidth]{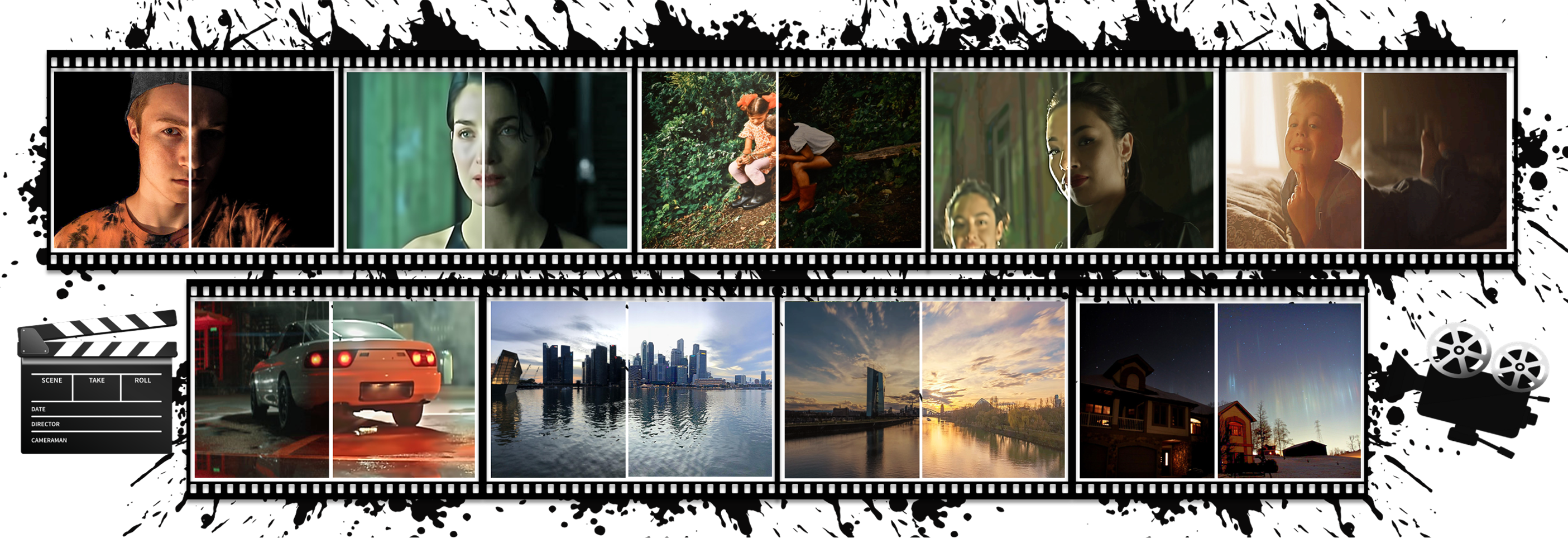}
	\caption{\textbf{Visual performance in real-world scenes with 4K resolution.} Our DPLUT achieves visually pleasing results in terms of brightness, color, contrast, and naturalness across diverse scenes and under various light distributions.}
 \label{teaser}
\end{figure*}


The lookup table (LUT) is a crucial component in the image signal processors (ISPs)~\cite{ISP} owing to its high efficiency and practicality. LUTs can be either pre-defined or learned. Pre-defined LUTs are manually tuned by experienced experts, exhibiting limited expressive capacity to adapt to diverse scenes. In contrast, learned LUTs~\cite{3DLUT,wang2021real,yang2022seplut} are significantly more expressive and have achieved promising outcomes in global exposure adjustment, contrast, and saturation, whereas challenges persist when applying LUTs for LIE.
\textit{\textbf{First}}, the contrast and pixel values of low-light images can be extremely small. Classic LUTs fail to adjust the local contrast of such images and the enhanced results may suffer from color shift and artifacts. This limitation aligns with the findings in~\cite{3DLUT,liu20234d}. \textit{\textbf{Second}}, current LUT-based image enhancement methods overlook the inherent sensor noise and artifacts concealed in low-light regions, which is also pointed out in ~\cite{3DLUT,conde2023nilut}.
\textit{\textbf{Third}}, the paradigm of learning LUTs in an unsupervised fashion remains challenging.


To tackle the aforelisted issues, we introduce a novel LIE framework termed DPLUT, which aims to achieve higher quality and efficiency simultaneously by taking advantage of lookup tables and diffusion priors. The proposed DPLUT consists of two key components: a light adjustment lookup table (LLUT) and a noise suppression lookup table (NLUT). Firstly, LLUT is crafted to generate coarse normal-light images. We treat LIE as a curve mapping issue, adopting LLUT to estimate pixel-wise curve parameters. By explicitly combining the image-specific curve function and LUT, we can effectively perform the mapping within a wide dynamic range, and ensure the enhanced image has a correct local contrast.
Secondly, to remove amplified noise and artifacts introduced from LLUT, NLUT is developed that marries the prior knowledge from the diffusion model to achieve real-time and high-quality image enhancement. \textit{\textbf{LLUT and NLUT are trained in an unsupervised manner. Notably, the diffusion model is only introduced in NLUT learning phase. With LLUT and NLUT, our solution can cope with diverse light distributions in real-time.}} The enhanced results are more clean and natural compared with existing state-of-the-art (SOTA) approaches. 

Our contributions can be summarized as follows:
\begin{itemize}

\item We develop a new unsupervised LIE framework with two lookup tables, i.e., a light adjustment lookup table and a noise suppression lookup table. 

\item We introduce diffusion priors and curve mappings to promote enhancement efficiency and effectiveness.

\item Extensive evaluations on three benchmark datasets show that DPLUT achieves state-of-the-art performance and can enhance 4K low-light images in real-time.

\end{itemize}

 \section{Related Work}
\subsection{Low-light Image Enhancement}
Enhancing images in low-light conditions has been a longstanding issue and great progress has been made over the decades. They can be roughly categorized into efficiency- and quality-oriented techniques. Efficiency-oriented approaches aim to construct lightweight enhancement models for mobile and source-limited platforms~\cite{cheng2004simple,huang2012efficient,abdullah2007dynamic,rahman2016adaptive}. For example, Wang et al.~\cite{wang2009real} enhanced the visibility and contrast via gamma correction and dynamic contrast ratio improvement. Guo et al.~\cite{Guo1} proposed to refine the initial estimated illumination map by imposing a structure prior.
Guo et al.~\cite{ZeroDCE} presented a reference-free LIE algorithm based on curve estimation, which can effectively perform mapping within a wide dynamic range. Ma et al.~\cite{SCI} established a cascaded illumination estimation process to achieve fast and robust LIE in complex scenarios.
Despite efficiency, the enhancement performance of current efficiency-oriented approaches is greatly inferior to quality-oriented methods~\cite{Lore,wang2022low,hou2023global, SNR,CUE,UHDFourICLR2023,Neco,liu2023nighthazeformer,zou2024vqcnir}. Lore et al.~\cite{Lore} designed a stacked sparse denoising auto-encoder to enhance low-light images. Lv et al.~\cite{MBLLEN} presented a multi-branch network that extracts rich features from different levels to enhance low-light images via multiple sub-networks. Xu et al.~\cite{SNR} incorporated the signal-to-noise ratio (SNR) prior to achieving spatial-varying LIE. Cai et al.~\cite{Retinexformer} designed a sophisticated transformer-based algorithm for LIE. 
Hou et al.~\cite{hou2023global} devised a diffusion-based framework and introduced a global structure-aware regularization to preserve the image's details and textures. Yi et al.~\cite{yi2023diff} combined the diffusion model with Retinex model for low-light image enhancement.
Quality-oriented approaches require deep and complex network structures and a huge amount of computational resources. As a classic prepossessing task, the practicality of quality-oriented methods is limited. 

\begin{figure*}[t!]
	\centering
 \setlength{\abovecaptionskip}{0.1cm} 
    \setlength{\belowcaptionskip}{-0.1cm}
	\includegraphics[width=1\linewidth]{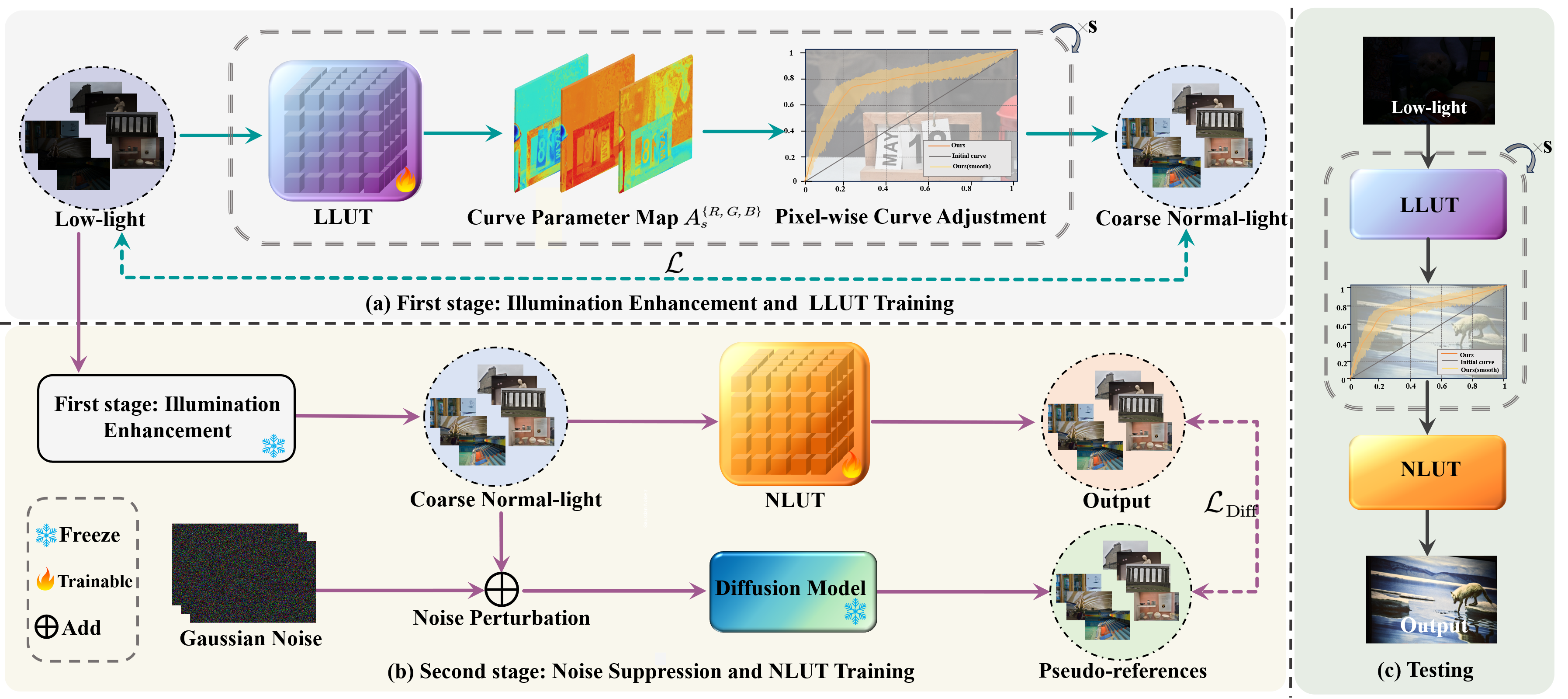}
	\caption{\textbf{The overall framework of our proposed DPLUT}. \textbf{In the training phase}, DPLUT involves two main stages. (a) In the first stage, we learn a light adjustment lookup table (LLUT), which estimates pixel-wise curve parameters for yielding coarse normal-light images. (b) In the second stage, we learn a noise suppression lookup table (NLUT) by introducing knowledge of a diffusion model, aiming at removing the amplified noise and artifacts introduced from LLUT. \textbf{In the testing phase}, with the LLUT and NLUT, DPLUT can robustly recover perceptual-friendly results in real-time.}
 \label{overview}
\end{figure*}

\subsection{LUTs for Image Enhancement}
The lookup tables (LUTs) are commonly used in ISPs~\cite{ISP,delbracio2021mobile,conde2022model}, especially some resource-constrained devices, to accelerate computation. The mapping procedure can be evaluated using only memory access and interpolation without performing the computation again. Due to its portability, various LUT based solutions have been proposed for photo enhancement~\cite{3DLUT,yang2022adaint,yang2022seplut,liu20234d}. For instance, 
Zeng et al.~\cite{3DLUT} first leveraged a lightweight CNN to predict the weights for integrating multiple basis LUTs, and the constructed image-adaptive LUT is utilized to enhance photos. Wang et al.~\cite{wang2021real} proposed a spatially-aware LUT that considers the global and local information. Liang et al.~\cite{liang2021ppr10k} improved the LUTs performance by adjusting learning strategies. Cong et al.~\cite{cong2022high} embedded the LUT-based sub-module in their network for efficient processing of high-resolution images.
Yang et al.~\cite{yang2022adaint} focused on improving the sampling strategy for 3D LUTs. Yang et al.~\cite{yang2022seplut} combined 1D-LUT and 3D-LUT to promote the enhancement performance while reducing computational costs.


\section{Preliminaries}
\textbf{3D-LUT.} 3D-LUT is a widely used image enhancement tool that maps the input color values to the corresponding output color values.
A classical 3D-LUT is defined as a 3D cube that contains $N^3$ elements, where $N$ is the number of bins in each color channel. Each element defines a pixel-to-pixel mapping $\mu^c(i, j, k)$, where $i, j, k=0,1, \ldots, N$-1 are elements' coordinates and $c$ indicates one of the channels. Given an input image $\left\{I_{(i, j, k)}^r, I_{(i, j, k)}^g, I_{(i, j, k)}^b\right\}$, the mapping procedure can be formulated as follows:
\begin{equation}
O_{(i, j, k)}^c=\mu^c\left(I_{(i, j, k)}^r, I_{(i, j, k)}^g, I_{(i, j, k)}^b\right),
\label{lut1}
\end{equation}
where $O^c$ is the output of 3D-LUT, $c \in\{r, g, b\}$, and $r, g, b$ is the color value of the red, green, blue channel, respectively. This mapping contains two basic operations, i.e., lookup and interpolation. 
The lookup operation is conducted to find its coordinates (i.e., $i, j, k$) in the 3D-LUT cube. Then, the output can be derived by the trilinear interpolation operation using its nearest eight surrounding elements. More detailed descriptions can be found in the supplementary material. Notably, as the value of $N$ increases, the 3D color transformation space becomes more accurate. Nevertheless, a large $N$ introduces massive parameters, leading to heavy memory burden, high training difficulty, and limited cell utilization. For simplicity, all LUTs mentioned in this paper refer to 3D-LUTs.

\textbf{Denoising Diffusion Model.} 
Diffusion model has shown remarkable promise in visual generation~\cite{guo2024versat2i,chai2023stablevideo,bai2023integrating,cao2023difffashion}, and it enlightens other tasks like image restoration~\cite{ye2024learning,ye2023adverse,ye2023sequential,chen2023sparse}, image fusion~\cite{lin2023domain,wang2023learning,wang2024cross,huang2023dp}, dehazing~\cite{ye2021perceiving,chen2023dehrformer}, and desnowing~\cite{chen2023cplformer,chen2022msp,chen2022snowformer}. Denoising diffusion models generate images by gradually denoising from a gaussian noise $p\left( x_T \right) =\mathcal{N}(0,\mathrm{I)}$ and transforming into a certain data distribution. The forward diffusion process $q\left( x_t\mid x_{t-1} \right)$ adds Gaussian noise to the image ${x}_t$. The marginal distribution can be written as: $q\left( x_t\mid x_0 \right) =\mathcal{N}\left( \alpha _tx_0,\sigma _{t}^{2}\mathrm{I} \right)$, where $\alpha_t$ and $\sigma_t$ are designed to converge to $\mathcal{N}(0,\mathrm{I)}$ when $t$ is at the end of the forward process ~\cite{kingma2021variational,song2020score}. The reverse diffusion process $p\left(\boldsymbol{x}_{t-1} \mid \boldsymbol{x}_t\right)$ learns to denoise. Given infinitesimal timesteps, the reverse diffusion process can be approximated with Gaussian ~\cite{song2020score} related with an optimal MSE denoiser ~\cite{sohl2015deep}. The diffusion models are designed as noise estimators $\boldsymbol{\epsilon}_\theta\left(\boldsymbol{x}_t, t\right)$ taking noisy images as input and estimating the noise. They are trained via optimizing the weighted evidence lower bound (ELBO) ~\cite{ho2020denoising,kingma2021variational}:
\begin{equation}
\mathcal{L}_{\mathrm{ELBO}}(\theta)=\mathbb{E}\left[w(t)\left\|\boldsymbol{\epsilon}_\theta\left(\alpha_t \boldsymbol{x}_0+\sigma_t \boldsymbol{\epsilon} ; t\right)-\boldsymbol{\epsilon}\right\|_2^2\right],
\end{equation}
where $\epsilon \sim \mathcal{N}(0,\mathrm{I)}$, $w(t)$ is a weighting function. In practice, setting $w(t) = 1$ delivers good performance ~\cite{ho2020denoising}. Sampling from a diffusion model can be either stochastic ~\cite{ho2020denoising} or deterministic ~\cite{song2020denoising}. After sampling $x_T\sim \mathcal{N}(0,\mathrm{I)}$, we can gradually reduce the noise level and reach a clean image with high quality at the end of the iterative process.

\section{Methodology}
The overall framework of our proposed DPLUT involves two main stages, as illustrated in Fig. \ref{overview}. In the first stage, we learn a light adjustment lookup table (LLUT) by a set of unsupervised losses, which maps the input RGB values to the corresponding pixel-wise curve parameter. With the LLUT, we can obtain coarse normal-light images. In the second stage, to remove amplified noise and artifacts introduced from LLUT, we learn a noise suppression lookup table (NLUT) through injecting knowledge of a diffusion model. It should be noted that the diffusion model and LLUT remain fixed during the training of NLUT. In the testing phase, with LLUT and NLUT, our solution can cope with diverse light distributions and achieve real-time and high-quality image enhancement. \textit{\textbf{Note that LLUT and NLUT are decoupled, i.e., LLUT can yield favorable results without NLUT.}} We provide further details on the key components of our approach below.

\begin{table}[!t]
\setlength{\abovecaptionskip}{0.1cm} 
\centering
\caption{Architecture of the LUT generator, where $k$ is a hyper-parameter that serves as a channel multiplier controlling the width of each convolutional layer. $N$ is the size (number of elements along each dimension) of the LUT.}\label{architecture}
\scalebox{0.78}{
\setlength\tabcolsep{6pt}
\renewcommand\arraystretch{1}
\begin{tabular}{ccc}
\rowcolor{mygray}
\hline ID  & Layer  & Output Shape  \\
\hline 0 & \text { Bilinear Resize } & $3 \times 256 \times 256$ \\
1 &  Depthwise Separable Conv3x3, LeakyReLU  & $k \times 128 \times 128$ \\
2 &  InstanceNorm  & $k\times 128 \times 128$ \\
3 &  Depthwise Separable Conv3x3, LeakyReLU  & $2k \times 64 \times 64 $\\
4 &  InstanceNorm  & $2k \times 64 \times 64$ \\
5 & Depthwise Separable Conv3x3, LeakyReLU  & $4k \times 32 \times 32$ \\
6 &  InstanceNorm  & $4k \times 32 \times 32$ \\
7 &  Depthwise Separable Conv3x3, LeakyReLU  & $8k \times 16 \times 16 $\\
8 &  InstanceNorm  & $8k \times 16 \times 16$ \\
9 & Depthwise Separable Conv3x3, LeakyReLU & $8k \times 8 \times 8 $\\
10 &  Dropout (0.5)  & $8k \times 8 \times 8$ \\
11 &  AveragePooling & $8k \times 2 \times 2$ \\
12 & Reshape  & $32k$ \\
13 & Fully Connected Layer  & $M$ \\
14 & Fully Connected Layer  & $3N^3$ \\
15 &  Reshape & $3\times N\times N\times N$\\
\hline
\end{tabular}}
\vspace{-0.1cm}
\end{table}
\subsection{Light Adjustment Lookup Table  }
\textit{\textbf{Motivation for applying LUTs to predict curve parameters.}} Existing LUTs mainly focus on RGB-to-RGB mapping. As a result, they always require a relatively large table size (33 or 64 points) to address the diversity of color ranges, leading to a heavy memory burden and high training difficulty. In contrast, \textit{\textbf{LLUT implements a more simple and effective mapping, i.e., RGB-to-Curve parameters, which requires a small table size (9 points)}}. The superiority of curve mapping is twofold: 1) It is monotonic, ensuring the preservation of contrast between adjacent pixels; 2) It is simple and differentiable, which benefits the gradient back-propagation process and can facilitate convergence.

\begin{figure}[!t]
    \centering
\setlength{\abovecaptionskip}{0.1cm} 
    \setlength{\belowcaptionskip}{-0.3cm}
    \includegraphics[width=0.96\linewidth]{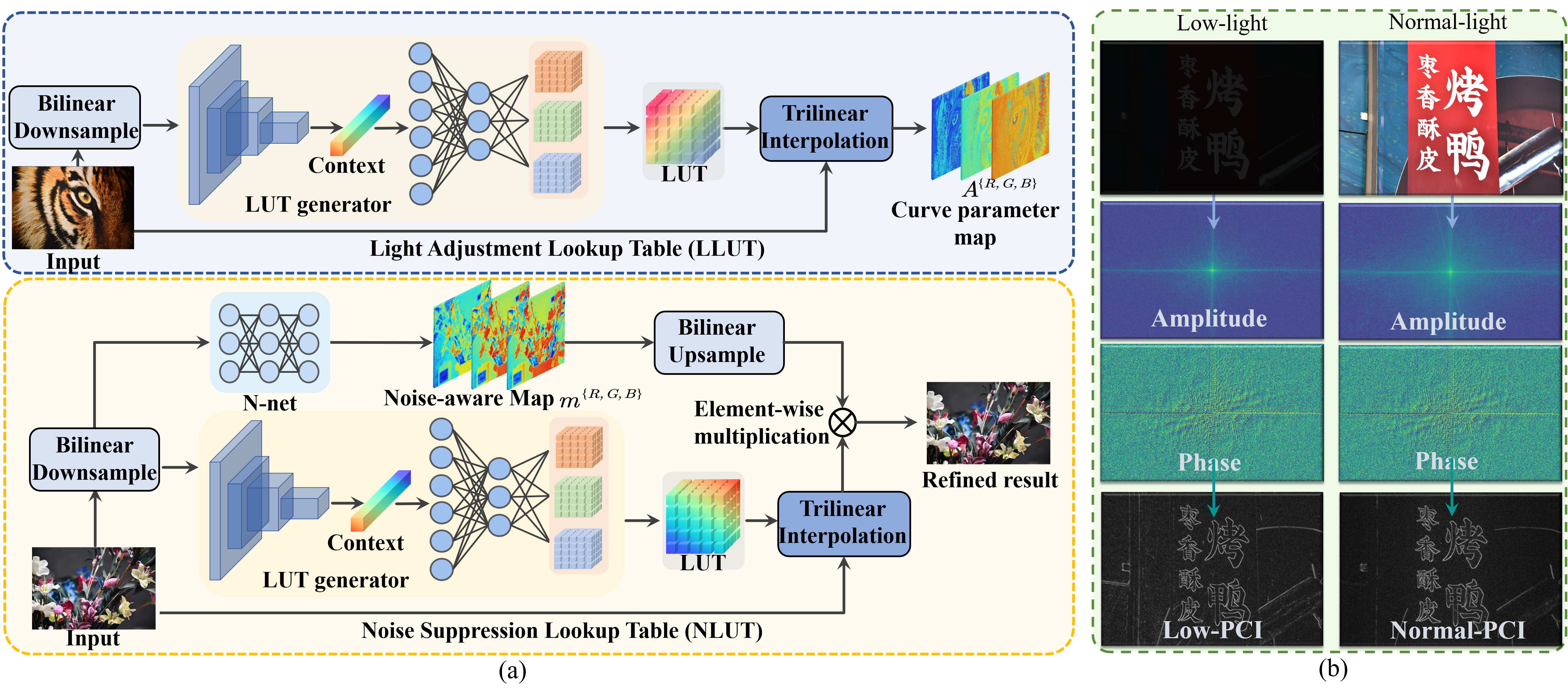}
    \caption{(a) The architecture of our two key components: a light adjustment lookup table (LLUT) and a noise suppression lookup table (NLUT). (b) We applied an inverse discrete Fourier transform to the phase of the low/normal-light image to obtain the phase-only reconstruction image (PCI) in the spatial domain. That means the amplitude of low/normal-light image is set to 1.}
    \label{conponents}
\end{figure}

As depicted in Fig. \ref{overview}(a), the first stage of our training framework involves the construction of a light adjustment lookup table (LLUT), which estimates the pixel-wise curve parameter for dynamic range adjustment of a specific image.
We begin by formulating our setting and introducing our notations. Given a low-light image $I\left(x \right) \in \mathbb{R}^{H\times W\times 3}$, LLUT maps the input RGB values to the corresponding curve parameter map ${A}(x)$. In order to automatically generate an image-adaptive LLUT, as shown in Fig. \ref{conponents}(a), we predict all the $N^3$ elements in the LUT by the neural network to consider the adaptation to the diversity of various input images. Such an objective formulates a mapping from the image context $D$ to a $3N^{3}$-dimension parameter space:
\begin{equation}
    \mathrm{LLUT}=f_{3D}(D),
\end{equation}
where $\mathrm{LLUT}$ is generated by the LUT generator module $f_{3D}\left( \cdot \right)$. The detailed architecture can be found in Tab.~\ref{architecture}.
Given the predicted LLUT, the estimated pixel-wise curve parameter map ${A}(x)$ for a specific image $I\left( x \right)$ can be formulated as:
\begin{equation}
A\left( x \right) =\,\,\mathrm{trilinear}\_\mathrm{interpolate} \left( \mathrm{LLUT},I\left( x \right) \right). 
\label{curve}
\end{equation}

Then, we recurrently apply the pixel-wise adjustment curve~\cite{ZeroDCE} to obtain the coarse normal-light sample $\hat{I}\left( x \right) \in \mathbb{R}^{H\times W\times 3}$. At the $s$-th step, the intermediate enhanced result $I_{s+1}(x)$ is:
\begin{equation}
I_{s+1}(x)=I_s(x)+\mathcal{A}_s(x)I_s(x)\left( 1-I_s(x) \right),
\label{LE}
\end{equation}
where $x$ denotes the pixel coordinates. The range of $A_s(x)$ is limited to between -1 and 1. Based on Eq.~\ref{LE}, to enable zero-reference learning in LLUT, the following four types of losses are adopted.

\textbf{Exposure loss.} This loss function ensures the enhanced image has a reasonable exposure level by penalizing the gray-scale intensity deviation from the mid-tone value:
\begin{equation}
\mathcal{L}_e=\frac{1}{z}\sum_{i=1}^z{\left\| V_i-v \right\| _{2}^{2}},
\end{equation}
where $z$ represents the number of non-overlapping local regions of size $16 \times 16$, and $V_i$ is the is the average intensity value of a local region in $\hat{I}$. We set $v=0.65$ in our experiments. 

\textbf{Structural consistency loss.} This loss function encourages spatial coherence of the enhanced image by minimizing phase error between the input image and its enhanced version:
\begin{equation}
\mathcal{L}_p=\left\| \mathcal{P}(I)-\mathcal{P}\left( \hat{I}\right) \right\| _1,
\end{equation}
where $\mathcal{P}(\cdot)$ indicates the phase in the Fourier domain. In Fig. \ref{conponents}(b), we perform the inverse discrete Fourier transform to obtain phase-only reconstruction images in the spatial domain. As observed, the phase-only reconstruction versions of low-light and normal-light exhibit structural consistency. This is because most illumination information is expressed as amplitudes, and structural information is revealed in phases. This conclusion is consistent with that in~\cite{UHDFourICLR2023}.

\textbf{Color loss.} This loss is based on the gray-world assumption, endeavoring to minimize the mean value difference between each color channel pair to correct the potential color deviations in the enhanced image:
\begin{equation}
\mathcal{L}_c=\sum_{(i,j)\in \xi}{\left( \hat{I}^{i}-\hat{I}^{j} \right) ^2},\xi \in \{(R,G),(G,B),(B,R)\}.
\end{equation}

\textbf{Smoothing loss.} This loss function is calculated in pixel-wise of each curve parameter map $A$, which preserves the monotonicity between neighboring pixels:
\begin{equation}
\mathcal{L}_s=\frac{1}{n}\sum_{s=1}^n{\sum_{c\in \delta}{\left( \left| \nabla _x\mathcal{A}_{s}^{c} \right|+\nabla _y\mathcal{A}_{s}^{c}\mid \right) ^2}},\delta =\{R,G,B\},
\end{equation}
where $n$ is the number of curve parameter maps. $\nabla_x$ and $\nabla_y$ represent the horizontal and vertical gradient operations, respectively.

The full objective function for LLUT is a weighted sum of all sub-loss terms:
\begin{equation}
\mathcal{L}\,\,=\,\,\lambda _1\mathcal{L}_e+\mathcal{L}_p+\lambda _2\mathcal{L}_c+\lambda _3\mathcal{L}_s,
\end{equation}
where $\lambda _1$, $\lambda _2$ and $\lambda _3$ are the weights of the losses, which are empirically set to 10, 5 and 1600 in all experiments.
\begin{table*}[!t]
\centering
\setlength{\abovecaptionskip}{0.1cm} 
\setlength{\belowcaptionskip}{-0.2cm}
\caption{Quantitative comparison on LOL, SICE and LSRW. “T”, “S”, and “U” represent “Traditional”, “Supervised” and “Unsupervised” methods, respectively. The best results of “S” and “U” are marked in \textcolor{blue}{blue} and \textcolor{red}{red}, respectively.}\label{table_results1}
\scalebox{0.8}{
\setlength\tabcolsep{8pt}
\renewcommand\arraystretch{1}
		\begin{tabular}{c|c|ccc|ccc|ccc|c}
			\toprule \rowcolor{mygray}
			~ & ~ & \multicolumn{3}{c|}{LOL} & \multicolumn{3}{c|}{SICE} & \multicolumn{3}{c|}{LSRW} & \\
			\cline{3-11} \rowcolor{mygray}	
			 \multirow{-2}*{Method} & \multirow{-2}*{Type} &
            PSNR$\uparrow$ & SSIM$\uparrow$ & LPIPS$\downarrow$ & PSNR$\uparrow$ & SSIM$\uparrow$ & LPIPS$\downarrow$ & PSNR$\uparrow$ & SSIM$\uparrow$ & LPIPS$\downarrow$ & \multirow{-2}*{Params (M)} \\
			\hline\hline
    SDD~\cite{SDD}& T     & 13.34 & 0.63  & 0.74  & 15.34 & 0.73  & 0.26  & 14.70 & 0.49  & 0.41 & -\\
    LECARM~\cite{LECARM} & T     & 14.40 & 0.54  & 0.32  & 18.59 & 0.78  & 0.26  & 15.33 & 0.42  & 0.32 & -\\
    \hline
    MBLLEN~\cite{MBLLEN}& S     & 15.25 & 0.70  & 0.32  & 18.41 & 0.73  & 0.31  & 16.87 & 0.51   & 0.45& 0.45\\
    RetinexNet~\cite{LOL} & S     & 17.60 & 0.64  & 0.38  &19.57  & 0.78  & 0.27  & 15.58 & 0.41  & 0.39 &0.84 \\
    DSLR~\cite{DSLR} & S     & 15.20 & 0.59  & 0.32  & 14.32 & 0.68  & 0.38  & 15.21 & 0.44  & 0.38 &14.93 \\
    DRBN~\cite{DRBN}& S     & 19.67 & 0.82  & 0.16  & 18.73 & 0.78  & 0.28  & 16.72 & 0.51  & 0.39&0.58 \\
    3DLUT~\cite{3DLUT}& S     & 16.36 & 0.64  & 0.35  & 15.53 & 0.64  & 0.38  & 15.74 & 0.48  & 0.43&0.59  \\
    Bread~\cite{Bread} & S     & 22.95 & 0.83  & 0.15  & 17.28 & 0.80  & 0.25  & 16.06 & 0.53  & 0.36&2.12 \\
    CUE~\cite{CUE} & S     & 22.67 & 0.79  & 0.20  & 20.06 & 0.82  & 0.24  & 18.19& 0.52  & 0.33 &\textcolor{blue}{0.26}\\
    DiffLL~\cite{jiang2023low} & S & \textcolor{blue}{26.19} & \textcolor{blue}{0.85}  & \textcolor{blue}{0.11}  & 21.33 & 0.84  & 0.22 &  \textcolor{blue}{19.27} &  \textcolor{blue}{0.55}  & \textcolor{blue}{0.30} &22.05\\
    Retinexformer~\cite{Retinexformer} & S     & 25.15 & 0.84  & 0.13 & \textcolor{blue}{22.32} & \textcolor{blue}{0.85}  & \textcolor{blue}{0.20}   &  19.23 &  0.54 & 0.31 &1.61\\
    \hline
    EnlightenGAN~\cite{EnlightenGAN}& U     & 17.48 & 0.65  & 0.32  & 18.73 & 0.82  & 0.23  & 17.05 & 0.46  & 0.33 &8.64 \\
    ZeroDCE~\cite{ZeroDCE}& U     & 14.86 & 0.55  & 0.33  & 18.67 & 0.80  & 0.26  & 15.84 & 0.45  & 0.31 &0.079 \\
    ZeroDCE++~\cite{Zerodcepp}& U     & 15.32 & 0.56  & 0.33  & 18.65 & 0.81  & 0.28  & 15.32 & 0.49  & 0.33&0.01 \\
    RUAS~\cite{RUAS} & U     & 16.40 & 0.49  & 0.27  & 13.21 & 0.72  & 0.43  & 14.31 & 0.48  & 0.47 &0.003 \\
    SCI~\cite{SCI} & U     & 14.78 & 0.52  & 0.33  & 15.94 & 0.78  & 0.51  & 15.24 & 0.42  & 0.45 & \textcolor{red}{0.0003}\\
    PairLIE~\cite{pairLIE}& U     & 19.46 & 0.73  & 0.24  & 21.23 & 0.83  & 0.22  & 17.59 & 0.49  & 0.32 &0.34\\
    NeRCo~\cite{Neco}& U   & 19.81 & 0.73  & 0.24  & 20.73 & 0.83  & 0.23  & 18.82 & 0.51  & 0.32 &23.30\\
    CLIP-LIT~\cite{clip-lie}& U     & 12.39 & 0.49  & 0.38  & 13.70 & 0.72  & 0.30  & 13.46 & 0.40  & 0.35&0.28 \\\hline
    \scalebox{1.5}{\color{sh_blue}{$\star$}} {\textbf{DPLUT (Ours)}} & U     & \textcolor{red}{20.66} & \textcolor{red}{0.74} & \textcolor{red}{0.22} & \textcolor{red}{21.27} & \textcolor{red}{0.84} & \textcolor{red}{0.21} & \textcolor{red}{18.91}& \textcolor{red}{0.53} & \textcolor{red}{0.28} &\textbf{0.078} \\ 
\bottomrule
\end{tabular}}
\end{table*}

\subsection{Noise Suppression Lookup Table }
LLUT learns the curve parameter mapping based on a set of unsupervised losses, its results might remain undesired noise and artifacts. Recently, the diffusion model has garnered considerable attention for its powerful generative capability and remarkable performance across various vision tasks. \textit{\textbf{In this paper, we introduce the powerful prior knowledge of the pre-trained diffusion model (PTDM) to facilitate noise suppression lookup table learning.}}
Specifically, as presented in Fig. \ref{overview}(b), we feed the coarse normal-light sample $\hat{I}$ to NLUT and PTDM, which generate final results $\hat{Y}$ and pseudo-references $Y$, respectively. As shown in Fig. \ref{conponents}(a), NLUT has a similar architecture to LLUT. The only difference is that we employ an additional lightweight network to estimate the pixel-wise noise-aware map for adjusting the output value. The output of NLUT can be formulated as:

\begin{equation}
\hat{Y}\left( x \right) =\,\,\mathrm{trilinear}\_\mathrm{interpolate}\left( \mathrm{NLUT},\hat{I}\left( x \right) \right) \odot m,
\end{equation}
where $\mathrm{NLUT}$ is generated by the LUT generator module $f_{3D}\left( \cdot \right) $, as shown in Tab.~\ref{architecture}.
$m=\left\{ m_{h,w} \mid h\in \right. \left. \mathbb{R}_{}^{H-1},w\in \mathbb{R}_{}^{W-1} \right\}$ is a noise-aware pixel-wise weight map for NLUT at location $(h, w)$, which is estimated by a lightweight network $\gamma \left( \cdot \right) $. Specifically, $\gamma \left( \cdot \right)$ contains six convolutional layers, and the first five layers are followed by a ReLU function to increase the nonlinear mapping ability.

Meanwhile, we use a PTDM to refine coarse normal-light sample $\hat{I}$ to pseudo-reference $Y$ through the forward and reverse steps. As illustrated in Fig. \ref{overview}(b), we first apply the diffusion forward process on $\hat{I}$ to sample $I_t$, which can be described as:
\begin{equation}
    q\left( I_t\mid \hat{I} \right) =\mathcal{N}\left( I_t;\sqrt{\bar{\alpha}_t}\hat{I},\left( 1-\bar{\alpha}_t \right) \hat{I} \right),
\end{equation}
where $t=0,1,...,T$-1, $T$ is the total number of iterations, and $I_t$ is the noisy image at time-step $t$. $\mathcal{N}$ represents the Gaussian distribution. $\bar{\alpha}_t=\prod_{t=0}^T \alpha_i$, where $\alpha_t=1-\beta_t$ and $\beta_t$ is the predefined scale factor. After obtaining $I_t$, the reverse process infers a noise-free sample $Y$ via iterative refinement, expressed as:
\begin{equation}
\begin{aligned}
I_{t-1}=&\sqrt{\bar{\alpha}_{t-1}}\left( \frac{I_t-\sqrt{1-\bar{\alpha}_t}\epsilon _{\theta}^{}\left( I_t \right)}{\sqrt{\bar{\alpha}_t}} \right)+\sqrt{1-\bar{\alpha}_{t-1}}\epsilon _{\theta}^{}\left( I_t \right)
\end{aligned},
\end{equation}
where $t=T,T$-1$,...,1$, $\epsilon _{\theta}\left( \cdot \right)$ is the noise estimator.
Note that, benefiting from the strong generative prior of the diffusion model, the recovered sample $Y$ exhibits less interference of noise and artifacts. Hence, we use $Y$ as the pseudo-reference to supervise the NLUT learning. The introduced diffusion prior can be expressed as:
\begin{equation}
\mathcal{L}_{Diff\,\,}=\left\| Y-\hat{Y} \right\| _1,
\end{equation}
where $\left\| \cdot \right\| _1$ is the $\ell_1$ regularization term. Notably, we only employ the diffusion model in the NLUT learning stage.

 \section{Experiments}
\subsection{Implementation Details}
All experiments are conducted on a single Titan RTX GPU, and the PyTorch framework is used to construct our networks. We employ an Adam optimizer with $\beta _1=\,\,0.9$ and $\beta _2=\,\,0.99$, batch size is set to 1. The training iterations of LLUT and NLTU are set to 200 and 300, respectively. The learning rates of LLUT and NLUT are $1e^{-4}$ and $1e^{-5}$, respectively.  The total number of curve steps for illumination enhancement is set to $n=8$. We utilize the pre-trained diffusion model on ImageNet~\cite{dhariwal2021diffusion} and employ the implicit sampling strategy (DDIM)~\cite{song2020denoising}. The total number of DDIM iteration steps is set to 100. We select the final 4 steps to implement the noise addition and removal process. The sizes of LLUT and NLUT are set to 9 and 17, respectively.

\subsection{Datasets} In order to validate the effectiveness of the proposed method, we use low-light images from LOL~\cite{LOL} and SICE-Part2~\cite{SICE} to train and test the network. The LOL dataset is officially divided into two parts, i.e., 485 low-light images for training and 15 low-light images for testing. SICE consists of 224 normal-light images and 783 low-light images. Each normal-light image corresponds to 2$\sim$4 low-light images. We use the first 50 normal-light images and corresponding 150 low-light images for testing and the rest 633 low-light images for training. For a more convincing comparison, we further extend evaluations on the LSRW dataset~\cite{LSRW}, which includes 1000 pairs for training and 50 ones for testing. 


\subsection{Comparison with the State-of-the-Art}
For a more comprehensive analysis, DPLUT is compared with 19 state-of-the-art LIE methods, which can be divided into the following three categories: traditional methods (SDD~\cite{SDD}, LECARM~\cite{LECARM}), supervised approaches (MBLLEN~\cite{MBLLEN}, RetinexNet~\cite{LOL}, DSLR~\cite{DSLR}, DRBN~\cite{DRBN}, 3DLUT~\cite{3DLUT}, Bread~\cite{Bread}, CUE~\cite{CUE}, DiffLL~\cite{jiang2023low}, Retinexformer~\cite{Retinexformer}), and unsupervised methods (EnlightenGAN~\cite{EnlightenGAN}, ZeroDCE~\cite{ZeroDCE}, ZeroDCE++~\cite{Zerodcepp}, RUAS~\cite{RUAS}, SCI~\cite{SCI}, PairLIE~\cite{pairLIE}, NeRCo~\cite{Neco}, and CLIP-LIT~\cite{clip-lie}). Note that the results of all those methods are reproduced by using the official codes with recommended parameters. 

\textbf{Quantitative Comparisons.} We employ three full-reference metrics, i.e., peak signal-to-noise ratio (PSNR) and structural similarity index (SSIM)~\cite{SSIM}, and learned perceptual image patch similarity (LPIPS)~\cite{LIPIS} to objectively evaluate the performance of each method. A higher PSNR/SSIM score indicates the result is closer to the reference. A lower LPIPS value denotes better enhancement performance. Tab. \ref{table_results1} reports the average assessment metrics on three datasets. The best results of supervised and unsupervised ones are highlighted in blue and red, respectively. The results show that DPLUT achieves the best performance among all unsupervised methods. Note that DPLUT performs on par with some supervised approaches, which demonstrates the effectiveness of our solution.
\begin{figure*}[t]
    \centering
\setlength{\abovecaptionskip}{0.1cm} 
    \setlength{\belowcaptionskip}{-0.2cm}
    \includegraphics[width=1\linewidth]{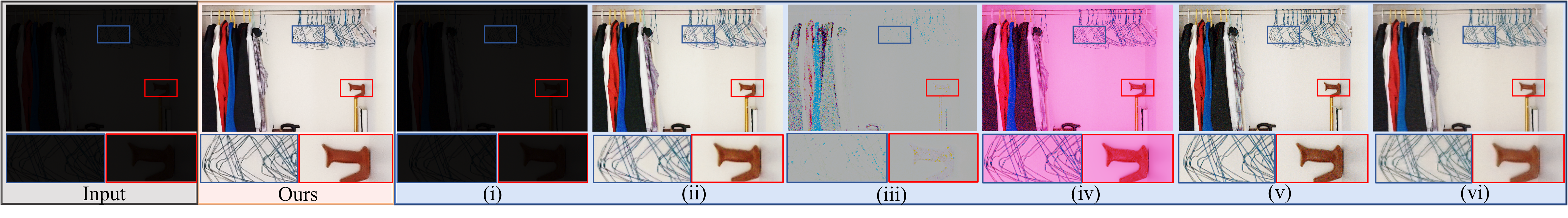}
    \caption{Visual comparisons of the ablation study. The full model achieves the best performance.}\label{ablation_vis}
\end{figure*}

\textbf{Visual Comparisons.}
For a more intuitive comparison, we further provide visual comparisons with other advanced algorithms in Figs \ref{LOL_result2}-\ref{mef_result2}. One can observe that DPLUT achieves visually pleasing results in terms of brightness, color, contrast, and naturalness. While other methods fail to cope with the extreme black light. Additionally, DPLUT can successfully suppress sensor noise in dark regions, and the result is clear and natural. In contrast, the competitors either amplify noise or are unable to correct color and contrast, leading to poor visual quality.

\begin{table}[!t]
\centering
\setlength{\abovecaptionskip}{0.1cm} 
\setlength{\belowcaptionskip}{-0.2cm}
\caption{Runtime (ms) comparison between our approach and SOTA methods on different resolutions. All methods are tested on one Titan RTX GPU. OOM means ``Out of Memory".}\label{runtime compare}
\scalebox{0.73}{
\setlength\tabcolsep{10pt}
\renewcommand\arraystretch{1}
\begin{tabular}{cccc}
\toprule
\rowcolor{mygray}
\ Resolution & $640 \times 480$ & $1920 \times 1080$ & $3840 \times 2160$ \\
\hline 
MBLLEN~\cite{MBLLEN} & 34.7 & 259.4 & 1045.1 \\
RetinexNet~\cite{LOL} & 20.4 & 139.1 & 550.4 \\
DRBN~\cite{DRBN}  & 15.3 & 127.5 & 550.4\\
3DLUT~\cite{3DLUT}  & 0.4 & 0.9 & 3.9 \\
Bread~\cite{Bread}  & 16.9 & 148.8 & 683.5 \\
CUE~\cite{CUE}& 31.3 & 228.5 & 1222.8 \\
DiffLL~\cite{jiang2023low}  & 152.7 & 1172.6 & 3579.1 \\
Retinexformer~\cite{Retinexformer}& 54.8 & 383.4 & OOM \\
EnlightenGAN~\cite{EnlightenGAN}& 5.1 & 38.9 & OOM \\
ZeroDCE~\cite{ZeroDCE} & 1.9 & 18.9 & 91.7 \\
ZeroDCE++~\cite{Zerodcepp} & 0.6 & 1.6 & 10.5 \\
RUAS~\cite{RUAS} &2.2 & 16.8 & 85.2 \\
SCI~\cite{SCI}& 0.4 & 1.4 & 10.1 \\
PairLIE~\cite{pairLIE}& 11.3 & 78.9 & 316.5 \\
NeRCo~\cite{Neco} &236.4 & OOM & OOM \\
CLIP-LIT~\cite{clip-lie} &9.6 & 86.1 & 347.6 \\
\scalebox{1.5}{\color{sh_blue}{$\star$}} {\textbf{DPLUT (Ours)}} &\textbf{ 6.3} &\textbf{ 6.6 }& \textbf{19.8} \\
\bottomrule
\end{tabular}
}
\vspace{-0.6cm}
\end{table}

\textbf{Inference Time Comparisons.}
Apart from the superior enhancement performance, another important advantage of our method is its efficiency. In this subsection, we report the inference time of three different resolutions, including 480P ($640\times 480$), 1080P ($1920\times 1080$), and 4K ($3840\times 2160$). For fair comparisons, we run all inference steps on a single Titan RTX GPU. Notably, we use the API call $torch.cuda.synchronize()$ to obtain precise feed-forward runtime. For each resolution, we record the average inference time on 100 images. In Tab. \ref{runtime compare}, as can be seen, DPLUT can handle all resolutions and the inference speed is considerably fast, especially for 4K images. EnlightenGAN~\cite{EnlightenGAN} and ZeroDCE~\cite{ZeroDCE} are faster than DPLUT at 480P resolution. However, as the image resolution increases, their inference speed decreases dramatically. Significantly, EnlightenGAN and ZeroDCE cannot handle 4K low-light images. Although 3DLUT~\cite{3DLUT}, SCI~\cite{SCI} and ZeroDCE++~\cite{Zerodcepp}) outperform our method in speed, their enhancement performance is inferior to DPLUT. The above analysis validates the superb performance and practicality of our approach.



\subsection{Ablation study}\label{ablation}
To understand the role of different components of our approach, we conduct several ablation studies on LOL~\cite{LOL}. 

\begin{figure}[t]
    \centering
\setlength{\abovecaptionskip}{0.1cm} 
    \includegraphics[width=1\linewidth]{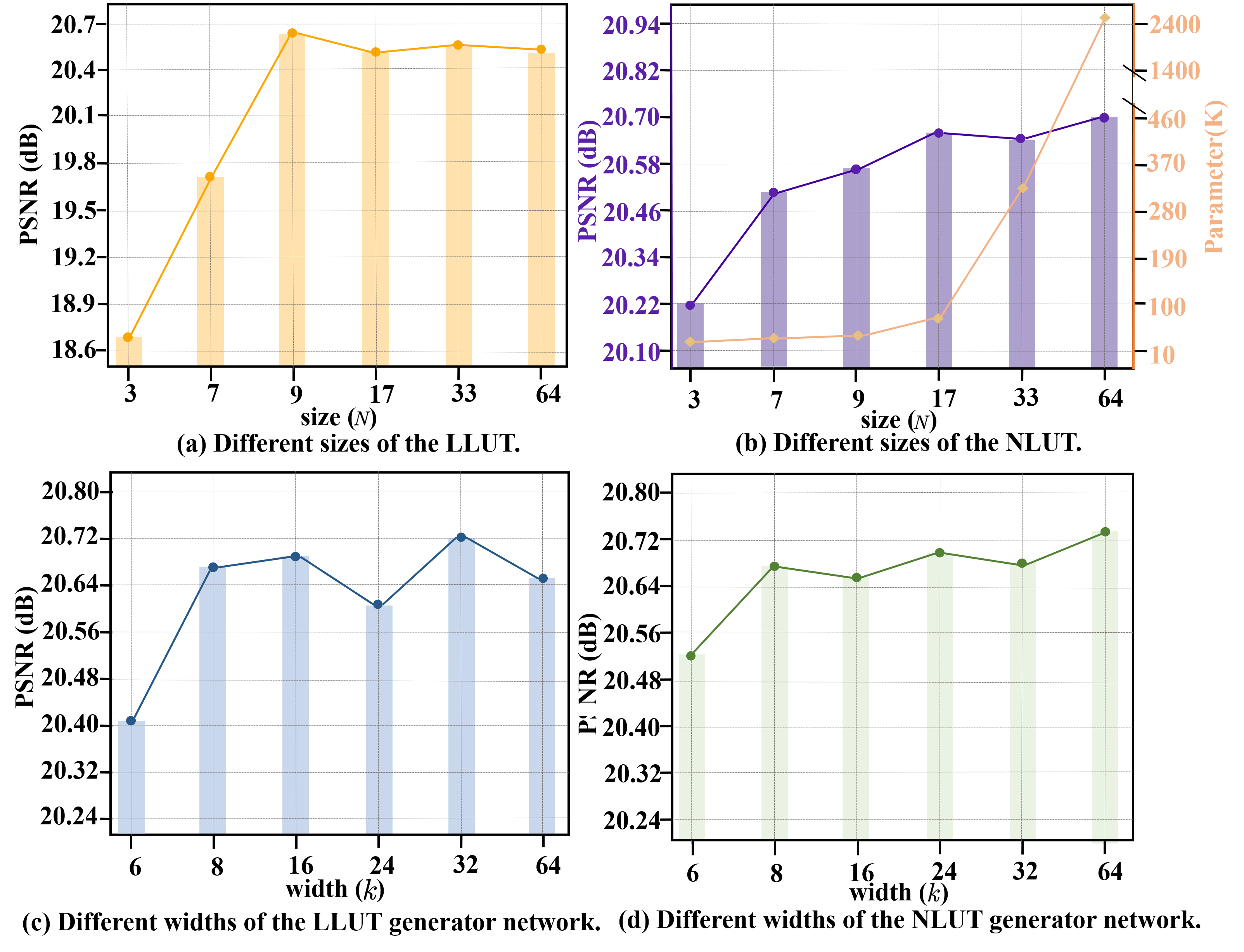}
    \caption{Ablation studies on different sizes of the LLUT and NLUT (a-b) and different widths of the LLUT and NLUT generator network (c-d).}
    \label{ablate_size}
    \vspace{-0.4cm}
\end{figure}

\textbf{Size of Lookup Tables.} 
(1) We first explore the influence of the size of LLUT. As shown in Fig. \ref{ablate_size}(a), increasing the size of LLUT will improve the performance continuously until the size is close to 9-point. Compared with existing learning-based LUTs (e.g., 33-point or 64-point LUTs~\cite{3DLUT,wang2021real,liu20234d}), the LUT size of our solution is relatively small. One potential explanation is that combining curve mapping and LUTs can promote the effectiveness of the mapping function.
(2) We investigate the impacts of the NLUT size. As shown in Fig. \ref{ablate_size}(b), enlarging the size of the NLUT yields limited performance gains but significantly increases the number of parameters, especially as the size exceeds 17. Such a phenomenon suggests the capacity redundancy of the NLUT. Consequently, the size of NLUT is set to 17 to better balance the performance and computational efficiency.

\textbf{Impact of the generator.} The generator of LUT is responsible for providing a coarse analysis of the input image. Here, we investigate the impact of the generator via ablating the network width, i.e., the hyper-parameter $k$. In Fig. \ref{ablate_size}(c-d), the ablation results indicate that increasing the width of the generator network does not always improve the performance. In contrast, it might increase capacity redundancy and training difficulties. Considering the trade-off between performance and memory footprint, we set $k=8$ for LLUT and NLUT. 

\begin{figure}[!t]
    \centering
\setlength{\abovecaptionskip}{0.1cm} 
    \setlength{\belowcaptionskip}{-0.5cm}
    \includegraphics[width=1\linewidth]{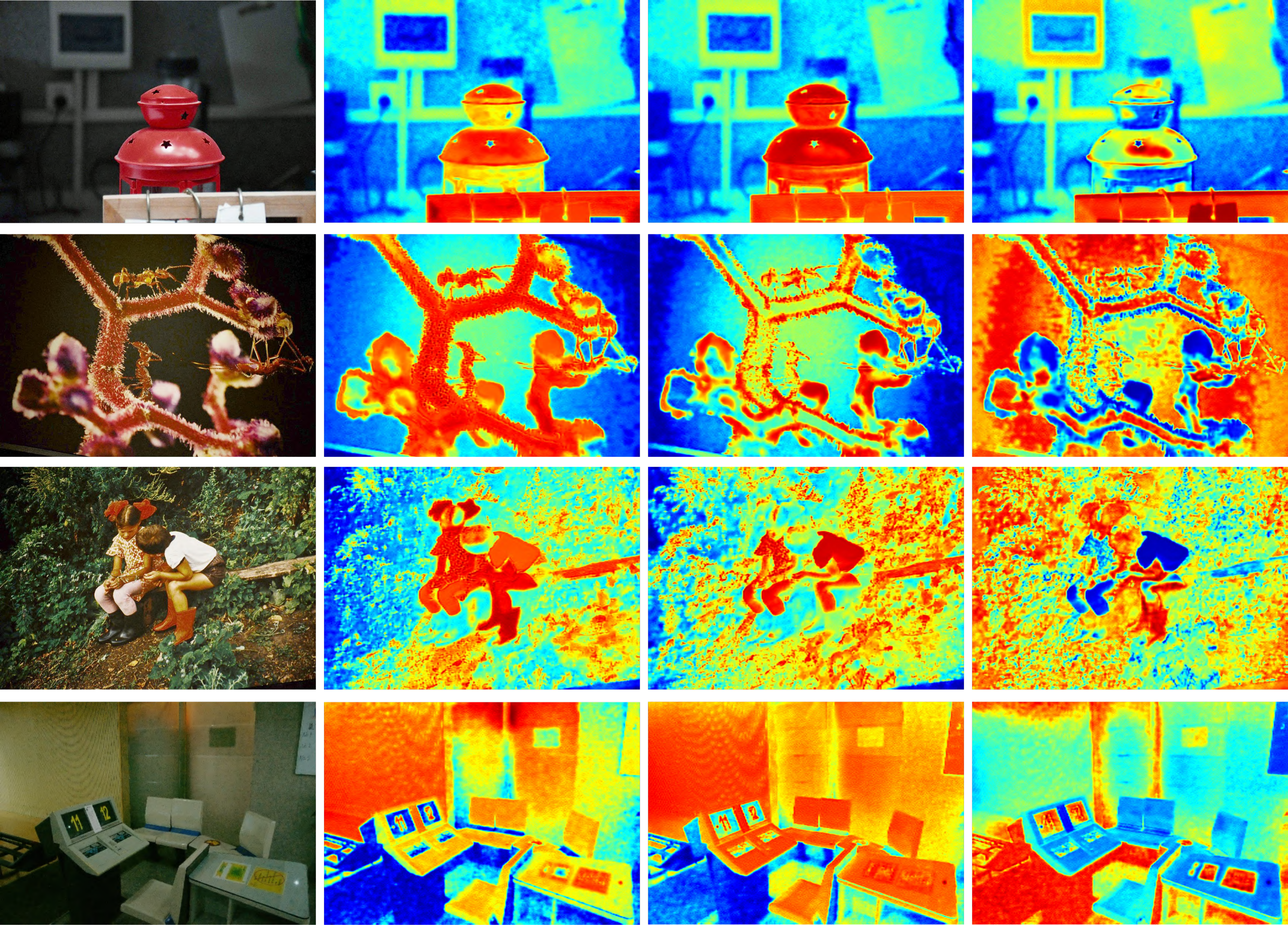}
    \caption{Visual results of the noise-aware weight map. In each row, the first is the coarse normal-light image, and the other three are visualizations for different channels. Red pixels indicate more activation, and blue pixels indicate less activation.}
    \label{noise_awar_map}
    \vspace{-0.1cm}
\end{figure}

\textbf{Effectiveness of the Noise-aware Map.} The noise-aware weight map is estimated by a lightweight predictor network. Its architecture is listed in supplementary material. We analyze the impact of the noise-aware weight map by adjusting the network width. As shown in Fig. \ref{ablate_size}(e), enlarging the width of the predictor network will improve its capability in noise removal. Meanwhile, the number of parameters also increases. These ablation results confirm the effectiveness of the noise-aware weight map, which can assist the NLUT in suppressing noise. Given the trade-off between memory footprint and denoising performance, the width of the predictor network is configured to 16. Furthermore, we also provide the visualization of the noise-aware weight map in Fig. \ref{noise_awar_map}.

\textbf{Impact of Iterations Steps.} We investigate the influence of the iteration steps of adding and removing noise. In Tab. \ref{ablation_iter}, we perform different numbers of iterations on the coarse enhanced samples to generate different versions of pseudo-references. One can see that fewer iteration steps lead to higher distortion metrics (i.e., PSNR and SSIM), whereas more iteration steps yield improved perception results (i.e., LPIPS) at the cost of increased processing time. This observation aligns with the findings in~\cite{whang2022deblurring}. By carefully balancing these metrics and the sampling time, we set the iteration steps as 4 to generate pseudo-references for the NLUT learning.

\textbf{Variants of the supervision.} We further conduct ablation studies to verify the effectiveness of the loss functions. Concretely, we have tested the following four variations over the original setting: (i) without the exposure loss. (ii) without the structural consistency loss. (iii) without the smoothing loss. (iv) without the color loss. (v) without the NLUT, i.e., only the LLUT. (vi) without prior knowledge of the pre-trained diffusion model and replace it with existing denoising method~\cite{MMBSN}.
Results are listed in Tab. \ref{ablation_loss} and Fig. \ref{ablation_vis}. We have the following observations: 1) removing the exposure loss fails to recover the low-light regions and the objective measures degrade significantly. 2) The structural consistency loss can promote the naturalness of the enhanced image. 3) Removing the smoothness loss hampers the correlations between neighboring regions, leading to obvious artifacts. 4) The absence of color loss results in serious color distortion. 5) Without the NLUT, i.e., only the LLUT, the enhanced image exhibits obvious noise and artifacts. In contrast, our full model generates clean and natural predictions, demonstrating the effectiveness of our approach. 6) Compared with the advanced diffusion model, training the NLUT with the existing denoising method~\cite{MMBSN} is inevitably constrained in terms of robustness and generalizability, yielding only suboptimal results as they either depend on synthetic datasets or necessitate hand-crafted assumptions. 

\begin{table}[!t]
\centering
\setlength{\abovecaptionskip}{0.1cm} 
\caption{Ablation study on different iteration numbers.}\label{ablation_iter}
\scalebox{1}{
\setlength\tabcolsep{2.5pt}
\renewcommand\arraystretch{1}
\begin{tabular}{c|cccc}
    \toprule 	
    \rowcolor{mygray}	
    Iterations & PSNR$\uparrow$ & SSIM$\uparrow$ & LPIPS$\downarrow$  & Time (s)\\
    \hline \hline			
    2 & 20.01 & 0.712 & 0.242&  0.764 \\
    3 & 20.21 & 0.733 & 0.231 & 1.095\\
    \textbf{4 }& \textbf{20.66} &\textbf{ 0.744} & \textbf{0.222}& \textbf{1.465}\\
    5 & 20.57 & 0.739 & 0.215&  1.795\\   
    10 & 20.19 & 0.712 & 0.211&  3.389\\   
    25 & 19.90 & 0.683 & 0.208&  8.709 \\   
    30 & 19.64 & 0.673 & 0.204&  10.482 \\   
    \bottomrule
\end{tabular}}
\end{table}
\begin{table}[!t]
\centering
\setlength{\abovecaptionskip}{0.1cm} 
\caption{Quantitative results of ablation studies on LOL.}\label{ablation_loss}
\scalebox{1}{
\setlength\tabcolsep{3pt}
\renewcommand\arraystretch{1}
\begin{tabular}{c|ccc}
    \toprule	
    \rowcolor{mygray}		
    Variants & PSNR$\uparrow$ & SSIM$\uparrow$ & LPIPS$\downarrow$ \\
    \hline \hline			
    (i) & 8.31 & 0.24 & 0.56 \\
    (ii) & 18.21 & 0.60 & 0.38 \\
    (iii) & 17.79 & 0.59 & 0.39\\
    (iv) & 17.49 & 0.56 & 0.40\\   
    (v) & 19.89 & 0.61 & 0.34\\   
    (vi) & 20.11 & 0.64 & 0.30\\   
    \midrule
   \scalebox{1.5}{\color{sh_blue}{$\star$}} {\textbf{DPLUT (Full)}}  & \textbf{20.66}  & \textbf{0.74} & \textbf{0.22}\\
    \bottomrule
\end{tabular}
}
\vspace{-0.4cm}
\end{table}

 \section{Conclusions}
In this paper, we introduce a novel unsupervised LIE framework based on lookup tables and diffusion priors (DPLUT) to achieve effective and efficient low-light image recovery. Two core components are devised to equip the framework, i.e., a light adjustment lookup table (LLUT) and a noise suppression lookup table (NLUT). Concretely, LLUT is designed to predict pixel-wise curve parameters for the dynamic range adjustment. NLUT is employed to remove the amplified noise after light brightening. Both LLUT and NLUT are trained in an unsupervised manner with a set of unsupervised losses and prior knowledge from a pretrained diffusion model, respectively. Extensive experimentation validates that our novel framework can enhance 4K low-light images in real-time and surpasses contemporary methods on three challenging benchmark datasets. In the future, we plan to optimize two lookup tables jointly to further promote the performance. Besides, we intend to apply our solution for different vision tasks.



{
   \small
   \bibliographystyle{ieeenat_fullname}
   \bibliography{main}
}
\clearpage


\begin{figure*}[!t]
    \centering
\setlength{\abovecaptionskip}{0.1cm} 
    \setlength{\belowcaptionskip}{-0.4cm}
    \includegraphics[width=1\linewidth]{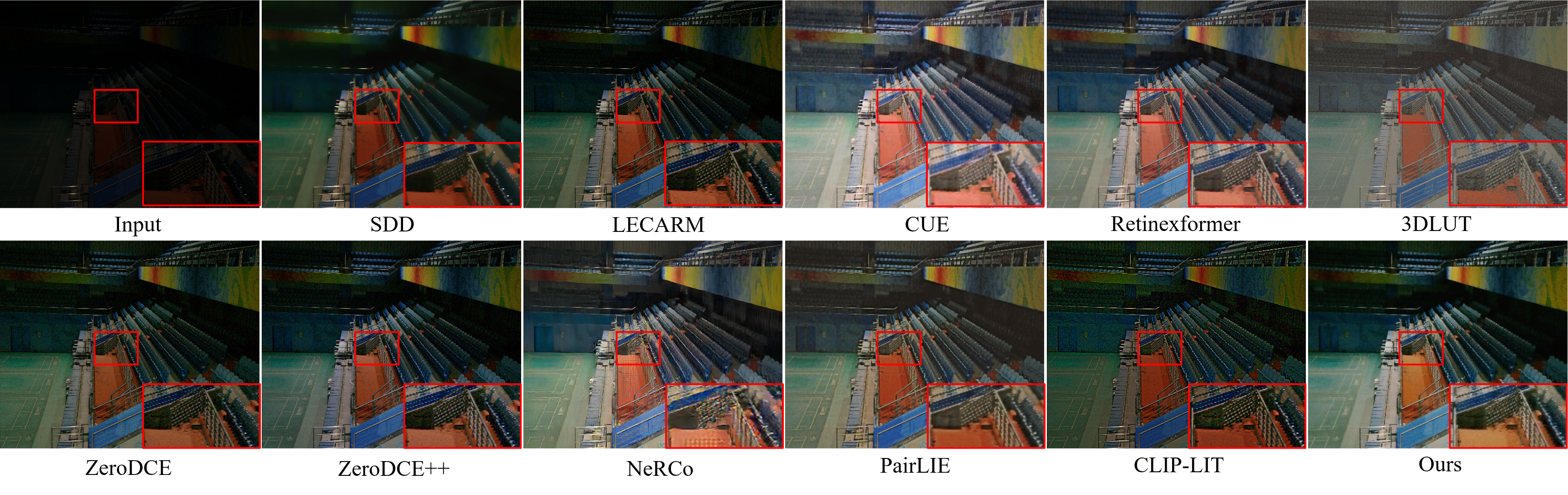}
    \caption{Qualitative comparison of our method and competitive methods on the LOL dataset.}\label{LOL_result2}
\end{figure*}

\begin{figure*}[!t]
    \centering
\setlength{\abovecaptionskip}{0.1cm} 
    \setlength{\belowcaptionskip}{-0.4cm}
    \includegraphics[width=1\linewidth]{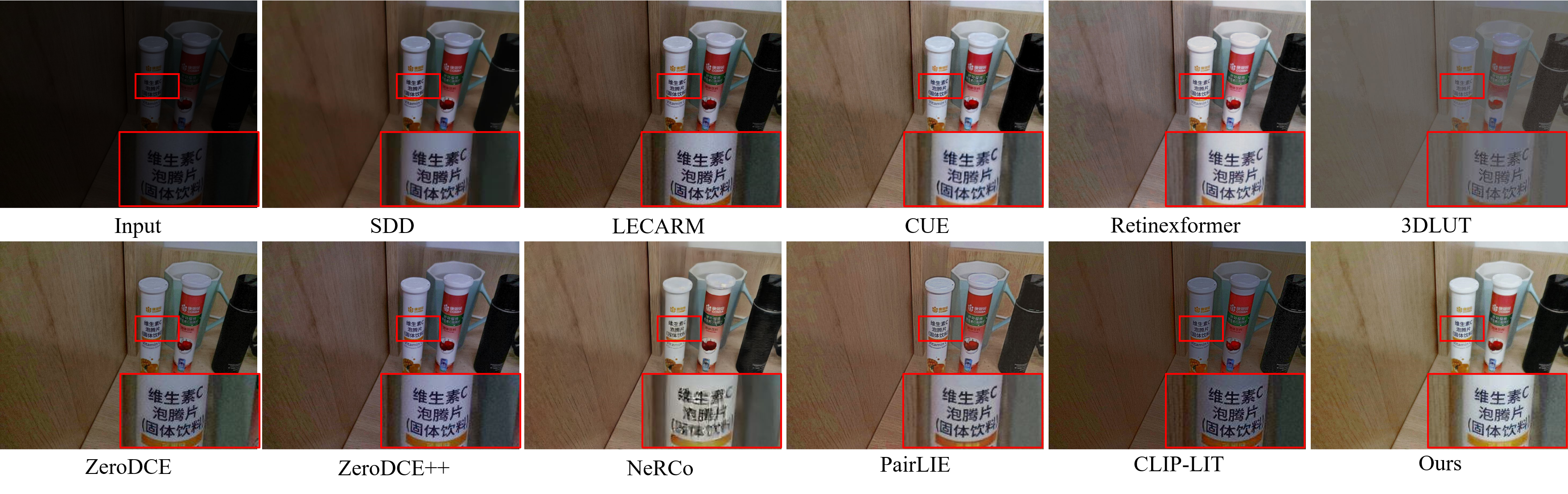}
    \caption{Qualitative comparison of our method and competitive methods on the LSRW dataset.}\label{lsrw2_result}
\end{figure*}

\begin{figure*}[!t]
    \centering
\setlength{\abovecaptionskip}{0.1cm} 
    \setlength{\belowcaptionskip}{-0.4cm}
    \includegraphics[width=1\linewidth]{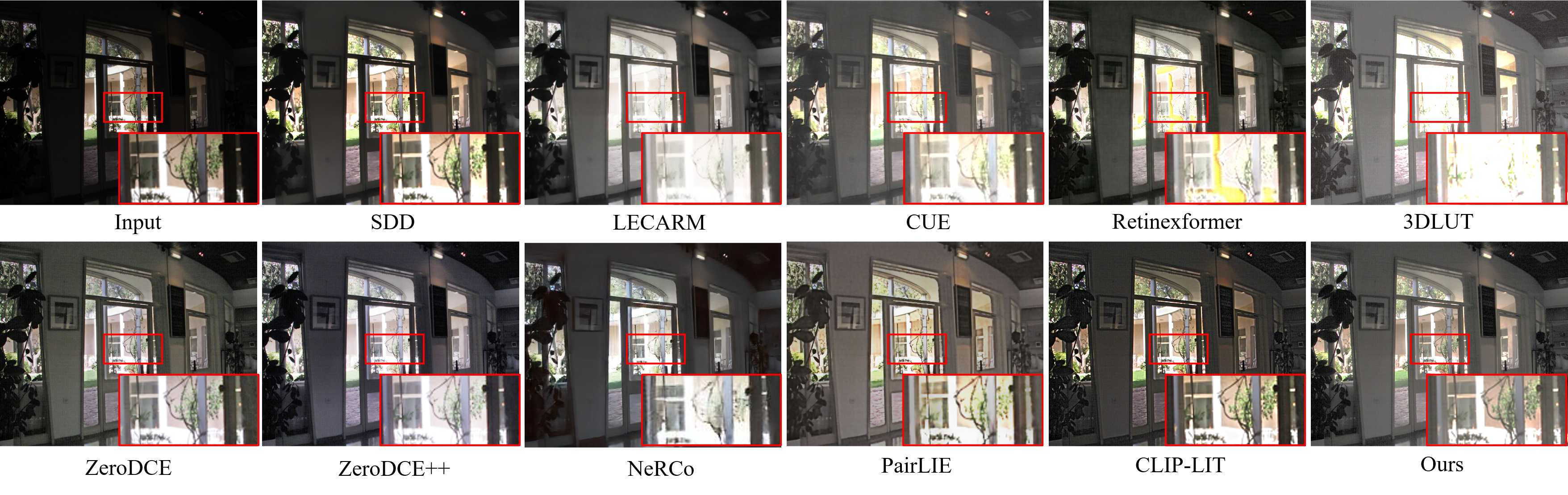}
    \caption{Qualitative comparison of our method and competitive methods on the MEF dataset.}\label{mef_result2}
\end{figure*}

\end{document}